\documentclass[runningheads]{llncs}

\usepackage{amsmath}
\usepackage{pgfplots}
\usepackage{makecell}
\usepackage{multirow}
\usepackage{graphicx}
\usepackage{booktabs}
\usepackage[colorlinks=true,linkcolor=blue,citecolor=blue,urlcolor=blue,]{hyperref}
\usepackage[doipre={doi:~}]{uri}
\usepackage{pgfplots}
\pgfplotsset{compat=1.8}

\usepackage{color}
\usepackage{CJKutf8}

\usepackage{marvosym}
\usepackage{ifsym}

\begin{document}
\begin{CJK}{UTF8}{gbsn}
\title{DialogRE$^{C+}$: An Extension of DialogRE to Investigate How Much Coreference Helps Relation Extraction in Dialogs}
\titlerunning{DialogRE$^{C+}$: Investigating How Much Coreference Helps DRE}
%
\author{Yiyun Xiong\inst{1} \and
Mengwei Dai\inst{1} \and
Fei Li\inst{1} 
Hao Fei\inst{2} \and
Bobo Li\inst{1} \and
Shengqiong Wu\inst{2} \and
Donghong Ji\inst{1} \and
Chong Teng\inst{1}$^{(\textrm{\Letter})}$}

\authorrunning{Y. Xiong et al.}

\institute{Key Laboratory of Aerospace Information Security and Trusted Computing, Ministry of Education, School of Cyber Science and Engineering, Wuhan University, Wuhan, China\\
\email{2018302180072@whu.edu.cn} \and
School of Computing, National University of Singapore, Singapore, Singapore}

\maketitle              

\begin{abstract}
Dialogue relation extraction (DRE) that identifies the relations between argument pairs in dialogue text, suffers much from the frequent occurrence of personal pronouns, or entity and speaker coreference.
This work introduces a new benchmark dataset DialogRE$^{C+}$, introducing coreference resolution into the DRE scenario.
With the aid of high-quality coreference knowledge, the reasoning of argument relations is expected to be enhanced.
In DialogRE$^{C+}$ dataset, we manually annotate total 5,068 coreference chains over 36,369 argument mentions based on the existing DialogRE data, where four different coreference chain types namely speaker chain, person chain, location chain and organization chain are explicitly marked.
We further develop 4 coreference-enhanced graph-based DRE models, which learn effective coreference representations for improving the DRE task.
We also train a coreference resolution model based on our annotations and evaluate the effect of automatically extracted coreference chains demonstrating the practicality of our dataset and its potential to other domains and tasks.

\end{abstract}

\section{Introduction}
\begin{figure}[!t]
\centering
\includegraphics[width=0.9\columnwidth]{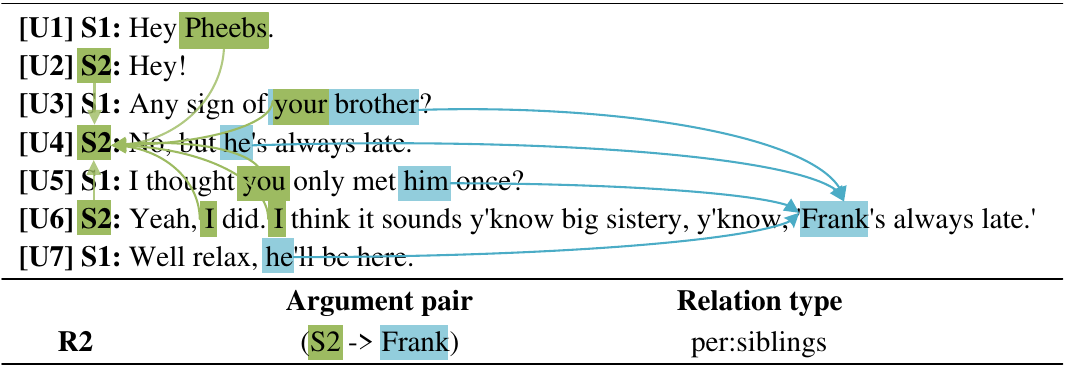}
\caption{An example dialogue and its corresponding relations in DialogRE \cite{yu2020dialogue}. 
S1, S2 denotes an anoymized speaker of each utterance ([U*]).
The words marked in green and blue are the mentions of subjective argument `S2' and objective argument `Frank'.}
\label{fig:dialog}
\vspace{-1.5em}
\end{figure}

Relation extraction (RE) is a long-standing task in natural language processing (NLP) community, aiming to detect semantic relationships between two arguments in one sentence \cite{zhou2016attention,fei2020boundaries}, or multiple sentences (i.e., document) \cite{yao2019docred,li-etal-2021-mrn,wang-etal-2022-entity}.
The latest research interest of RE has been shifted from the sentence level to the dialogue level, which identifies the relations between pairs of arguments in the conversation context.
As can be exemplified in Figure~\ref{fig:dialog}, DRE often comes with cross-sentence relations due to the characteristic of multiple parties and threads within dialogue context \cite{yu2020dialogue}.
This makes DRE much more challenging compared with the common sentence-level RE \cite{xue2022corefdre}

The primary challenge in DRE that causes difficulty to understand dialogue semantics lies in the \textbf{speaker coreference problem} \cite{zhou2021relation}.
In the dialogue scenario, speakers often use pronouns (e.g., `he', `she', `it') for referring to certain targets, instead of the target names, such as person, location and organization names.
Especially in DialogRE \cite{yu2020dialogue}, personal pronouns are frequently witnessed, which greatly hinder the relation reasoning of the task. 
As illustrated in Figure~\ref{fig:dialog}, Speaker2 (S2) are coreferred with various pronouns in different utterances (e.g., `you' in [U5], `I' in [U6]) refering to one identical person `Pheebs';
another person `Frank' are referred as `your brother' in [U3] and `him' in [U5] by different speakers.
Without correctly reasoning the coreference of speakers or their roles, it will be problematic to understand or infer the relations between arguments.
Unfortunately, previous DRE works either ignore the coreference information \cite{lee2021graph,qiu2021socaog} or utilize inaccurate or incomplete coreference information extracted based on heuristic rules \cite{long2021consistent,zhou2021relation,FeiDiaREIJCAI22}.

To this end, this work contributes to DRE with a new dataset \textbf{DialogRE$^{C+}$}, where all the coreference chains \cite{pradhan2012conll} are annotated manually by 11 graduate students and based on the existing DialogRE data.
To facilitate the utility of DialogRE$^{C+}$, we define four types of coreference chains, including \textit{Speaker Chain}, \textit{Person Chain}, \textit{Location Chain} and \textit{Organization Chain}.
Finally, DialogRE$^{C+}$ marks 36,369 mentions involved in total 5,068 coreference chains.

Based on the DialogRE$^{C+}$ dataset, we develop 4 coreference-enhanced graph-based DRE models, in which the coreference features are properly modeled and represented for learning comprehensive representations of arguments and better reasoning of the argument relations.
In addition, in order to explore the improvement effect of automatically extracted coreference chains on DRE, we train a coreference resolution model \cite{lee2017end} using the English coreference resolution data from the CoNLL-2012 shared task \cite{pradhan2012conll} and our DialogRE$^{C+}$, and then employ extracted coreference information in the DRE models.

Experimental results show that the inclusion of coreference chains in our DialogRE$^{C+}$ dataset has substantially enhanced the performance of each model, compared to their original counterparts
\cite{lee2021graph,zhou2021relation,zeng2020double,chen2023dialogue}. 
Specifically, this improvement is reflected in average F1 score increases of 2.8\% and 3.2\% on the development and test sets, respectively.
Moreover, the automatically extracted coreference chains improve 1.0\% and 0.6\% F1s on average compared with original models.
Further analysis demonstrates that the method, when augmented with annotated coreference information, exhibits superior performance in detecting cross-utterance relations.
We release the DialogRE$^{C+}$ dataset and the benchmark models to facilitate subsequent research.\footnote{\url{https://github.com/palm2333/DialogRE_coreference}}\footnote{\url{https://gitee.com/yyxiong715/DialogRE_coreference}}

\vspace{-4mm}
\section{Related Work}

\vspace{-2mm}
\subsection{Relation Extraction}

\vspace{-2mm}
\subsubsection{Intra-/Inter-Sentence RE.}
Relation extraction is one of the key tracks of information extraction \cite{Li00WZTJL22,FeiLasuieNIPS22,fei-etal-2023-constructing,cao-etal-2022-oneee}.
Most of the previous RE research focus on the sentence-level relation extraction, predict the relationships between two entities within a single sentence with neural network modeling \cite{zhou2016attention,bioinformatics-btaa993}.
Due to the fact that a large number of relations are expressed in multiple sentences in practice, the extraction scope has been expanded to the inter-sentence scenario. 
Nan et al. \cite{nan2020reasoning} empowered the relational reasoning across sentences by automatically inducing the latent document-level graph, and develop a refinement strategy to incrementally aggregate relevant information.
Tang et al. \cite{tang2020hin} proposed a hierarchical inference network by considering information from entity, sentence, and document levels.
GAIN \cite{zeng2020double} constructs two graphs to capture complex interactions among different mentions underlying the same entities.

\vspace{-3mm}
\subsubsection{Dialogue-Level RE.}
In 2020, Yu et al. \cite{yu2020dialogue} propose dialogue-based relation extraction dataset (DialogRE) and the DRE task on this basis.
REDialog \cite{zhou2021relation} designs speaker embeddings and speaker-pronoun coreference particularly for the features of dialogue text, and word-relation attention and graph reasoning are used to further enhance the model. 
TUCORE-GCN \cite{lee2021graph} is an utterance context-aware graph convolutional network. A heterogeneous dialogue graph is introduced to model the interaction between arguments in the whole dialogue.
HGAT \cite{chen2023dialogue} presents a graph attention network-based method where a graph that contains meaningfully connected speaker, entity, type, and utterance nodes is constructed.
Fei et al. \cite{FeiDiaREIJCAI22} construct dialogue-level mixed dependency graph (D$^2$G) for DRE with various conversational structure features, such as including dialogue answering structure, speaker coreference structure, syntactic dependency structure, and speaker-predicate structure.
D$^2$G has verified the explicit integration of speaker coreference helps MRE.
However, these models do not fully utilize coreference information,  which limits their ability to infer argument relations.

\vspace{-3mm}
\subsection{Applications of Coreference Resolution}
Coreference resolution is a core linguistic task that aims to find all expressions which refer to the same entity. 
Lee et al. \cite{lee2017end} proposed the first end-to-end neural coreference resolution system.
Based on this, many transformer-based models \cite{joshi2019bert,kirstain2021coreference} achieved remarkable success on the CoNLL benchmark \cite{pradhan2012conll}.
In recent years, long text-based tasks have become increasingly abundant. Researchers have noticed the importance of coreference information and applied coreference resolution to many downstream tasks.

Wu et al. \cite{wu2021coreference} developed an effective way to use naturally occurring coreference phenomena from existing coreference resolution datasets when training machine reading comprehension models.
Xue et al. \cite{xue2022corefdre} imitated the reading process of humans by leveraging coreference information when dynamically constructing a heterogeneous graph to enhance semantic information.
Coreference resolution is also applied in machine translation \cite{loaiciga2013anaphora}, summarization \cite{steinberger2007two}, dialogue \cite{strube2003machine} to improve the performance of the task.
In this work, we incorporate the coreference information obtained from a coreference resolution model into DRE models to explore its impact on the DRE task.

\vspace{-4mm}
\section{DialogRE$^{C+}$: An Extension of DialogRE with Coreference Annotations}

\vspace{-2mm}
\subsection{Annotation Method}

\vspace{-2mm}
In the annotation process, we employ the Brat annotation tool\footnote{http://brat.nlplab.org}, which is widely applied to annotate events, entities, relationships, attributes, etc.
Given a dialogue text, we annotate the pronouns that refer to the same argument or concept with coreference relations, including personal pronouns, possessive pronouns, and names.
We predefine four types of coreference chains: \textit{Speaker Chain}, \textit{Person Chain}, \textit{Location Chain}, and \textit{Organization Chain}.

\begin{figure*}[!t]
\centering
\includegraphics[width=0.9\textwidth]{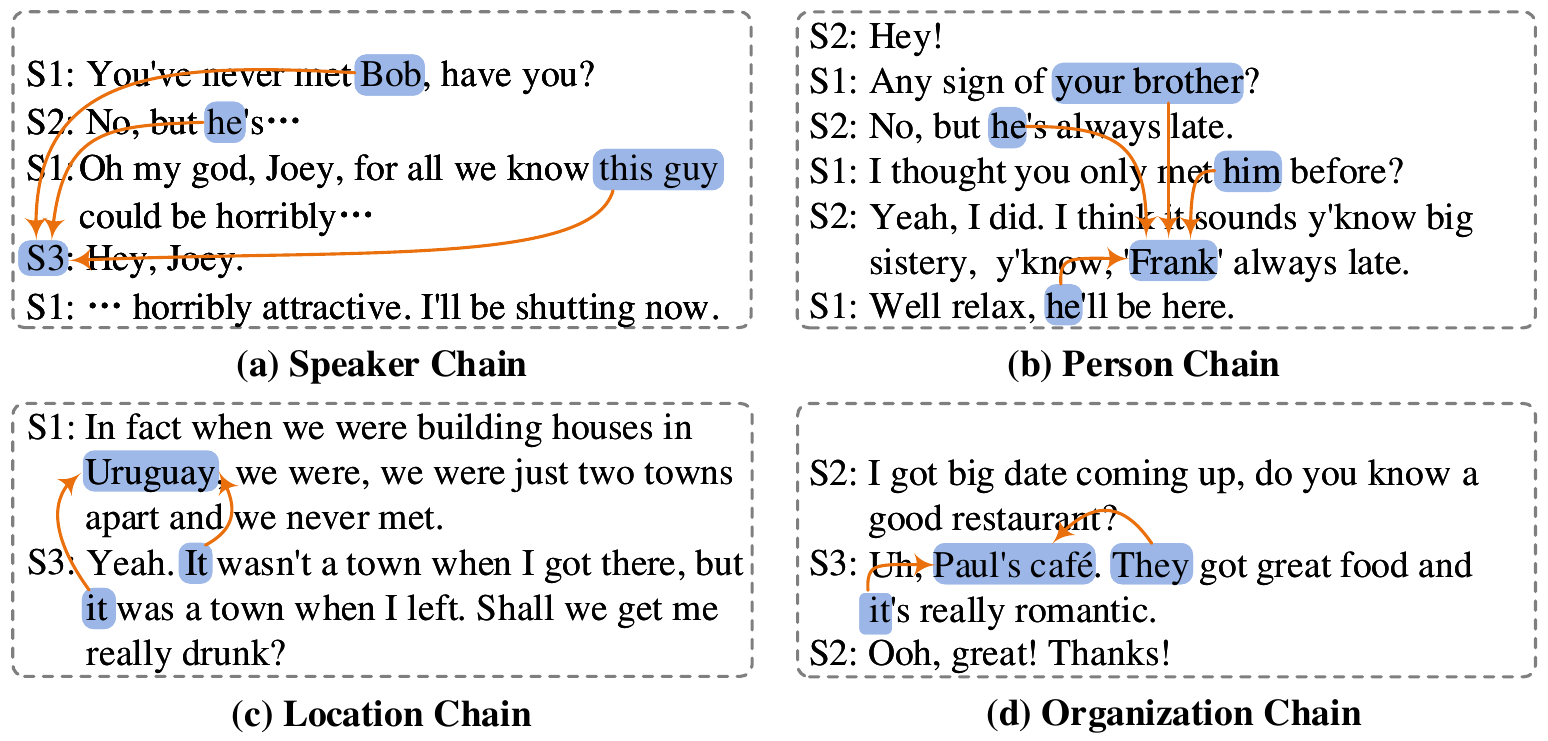}
\caption{The illustration of four types of coreference chains.}
\label{fig:exam}
\vspace{-1em}
\end{figure*}

\textbf{Speaker Chain} refers to the coreference chain of people who are currently involved in communication\footnote{Plural personal pronouns such as ``we, us, them and they'' refer to multiple entities or speakers, thereby they involve relationship extraction of multiple entities. 
In this paper, since focus on relationship extraction of two entities, we do not mark them for the time being.
}.
We focus on annotating the personal pronouns and names which refer to the same speaker when communicating, such as `I', `you', `he', `she', etc.
As shown in Figure~\ref{fig:exam}.(a), we notice that the pronouns, `Bob', `he', `this guy', refers to the same speaker, S3.
Thus, the final speaker chain built is $\left[S3 \gets(Bob, he, this\ guy)\right]$.

\textbf{Person Chain} refers to the coreference chain of people who are not currently involved in communication.
It is common that people discuss another person who does not appear in the dialogue, i.e., a third entity.
Person chain marks all pronouns and names referring to the same third entity.
For instance, in Figure~\ref{fig:exam}.(b), we find that the two speakers, S1 and S2, are talking about `Frank' who does not appear in the conversation.
Thus, we mark the person chain of `Frank' as $\left[Frank \gets(your\ brother, he, him, he)\right]$\footnote{The two pronouns, `he', appear in the third and sixth utterance, respectively.}.

\textbf{Location Chain} refers to the coreference chain of location, in which we annotate all names and pronouns of a place, such as ``it'', ``this'', etc.
In Figure~\ref{fig:exam}.(c), S1 mentioned a place, `Uruguay', and then S3 use the pronoun, `it', to represent the aforementioned place.
Finally, the marked location chain is $\left[Uruguay \gets(it, it)\right]$.

\textbf{Organization Chain} represents the coreference chain of organization, in which we annotate all names and referential pronouns of an organization that is discussed in a dialogue.
In Figure~\ref{fig:exam}.(d), `Paul's\ Café' is a new organization.
In the following, `They' and `it' both refer to `Paul's\ Café'.
Thus, the organization coreference chain in this dialogue is $\left[Paul's\ \textit{Café} \gets (They, it)\right]$.

\vspace{-3mm}
\subsection{Annotation Quality Control}
Before annotating, we have designed detailed annotation instructions. 
Then 11 graduate students are employed to annotate all the data.
After data annotation is completed, a senior annotator has examined all annotated data. 
If there is any contradiction, the senior annotator will discuss it with the corresponding annotator and reach a consensus.
In this way, we ensure that at least two annotators agree on each annotation result, achieving annotation consistency.

\begin{table}[t]
\centering
\scriptsize
\caption{The statistics for DialogRE$^{C+}$.
}
\setlength{\tabcolsep}{2.0mm}
\resizebox{0.6\columnwidth}{!}{
    \begin{tabular}{lrrr}
    \hline
    & \textbf{Train}  & \textbf{Dev}   & \textbf{Test}     \\ 
    \hline
    Speaker Chain      & 2,277 & 748  & 784  \\
    Person Chain       & 645   & 225  & 232   \\
    Location Chain     & 48    & 20   & 37       \\
    Organization Chain & 26    & 8    & 18      \\
    \hline
    Mentions        & 21,990 & 7,183  & 7,196  \\
    Coref. Chains & 2,996  & 1,001  & 1,071  \\ 
    \hline
    Dialogues & 1,073 & 358 & 357 \\
    Utterances & 14,024 & 4,685 & 4,420 \\
    Argument pairs & 5,997 & 1,914 & 1,862 \\
    \hline
    \end{tabular}
}
\label{statistic}
\end{table}

\vspace{-2mm}
\subsection{Data Statistics}
We annotate coreference chains based on the DialogRE dataset, which has a total of 36 relation types, 1,788 dialogues, 23,129 utterances, and 9,773 argument pairs, as shown in Table~\ref{statistic}.
In addition, we totally annotated 5,068 coreference chains and 36,369 mentions.
In other words, each chain contains about 7 (36369/5068) mentions.

\vspace{-3mm}
\section{Backbone Models for DRE and Coreference Resolution}

\begin{figure*}[t]
    \centering
    \includegraphics[width=1.0\textwidth]{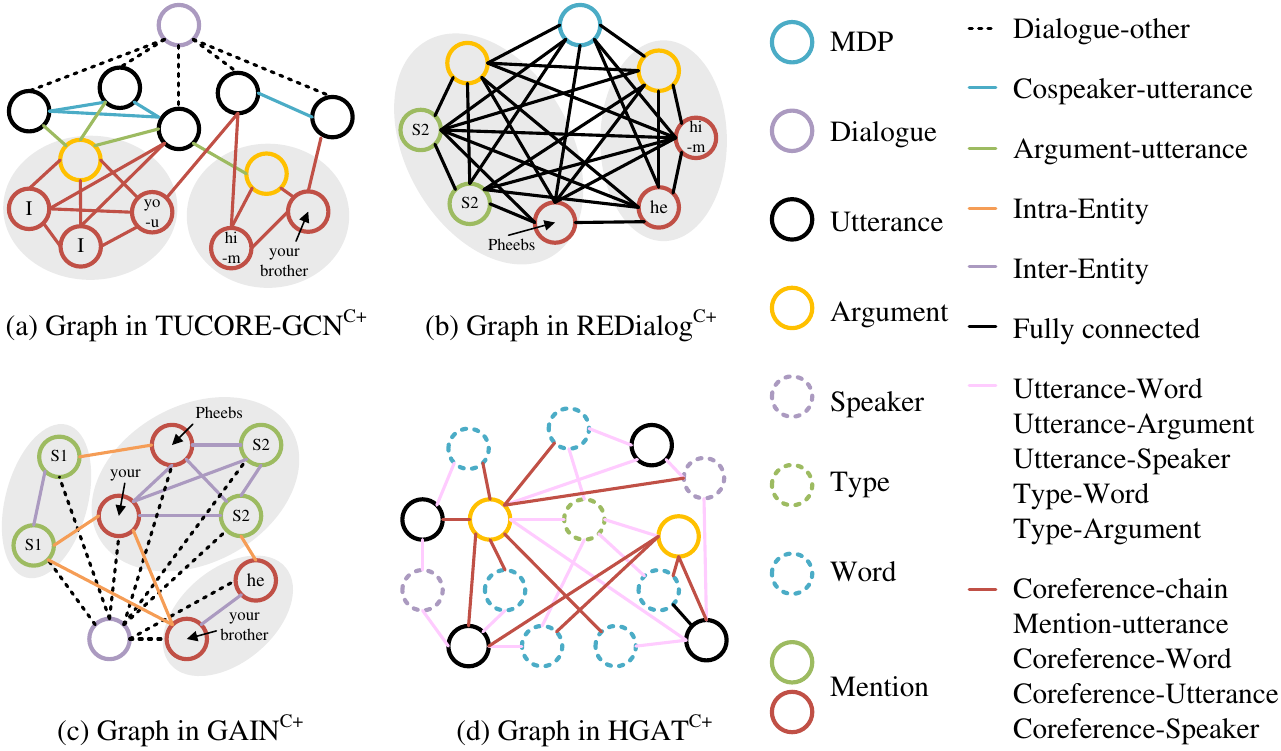}
    \caption{The coreference-enhanced graph in backbone models for the example in Figure~\ref{fig:dialog}. 
    Arguments are S2 ``Pheebs'' and her brother ``Frank''.
    The explanations of nodes and edges can be found in the paper of backbone models, except for the red ones, which are only used in this paper because they depend on coreference information. The gray backgrounds denote the coreference clusters containing argument entities and their mentions.
    } 
    \label{fig:DRE_graph}
\vspace{-1em}
\end{figure*}

\vspace{-2mm}
Graph-based models are widely used in the DRE task, as they can structure complicated syntactic and semantic relations. 
Therefore, we choose four graph-based models and enrich the dialogue graph using coreference chains to investigate how much coreference helps RE. 
Figure~\ref{fig:DRE_graph} shows the graph structure of these DRE models.\footnote{We do not compare with D$^2$G \cite{FeiDiaREIJCAI22} as it uses many other structural features than coreference information (e.g., dependency tree and dialogue answering links), which may cause unfair comparisons.}
It is worth noting that the red nodes are the nodes we propose or newly added mention nodes, and the red edges are our newly proposed edges.
In order to verify the effect of automatically extracted coreference information, we train a coreference resolution model, and add the machine-predicted coreference chains to the above graph-based models.

\vspace{-3mm}
\subsection{DRE Models}

\subsubsection{TUCORE-GCN$^{C+}$}

encodes the dialogue text with BERT$_s$ \cite{yu2020dialogue}, and then applies a masked multi-head self-attention to effectively extract the contextualized representation of each utterance from BERT$_s$.
Next, as shown in Figure~\ref{fig:DRE_graph}.(a), we build a heterogeneous graph over the pre-defined nodes and edges.
Thereafter, GCN is adopted to model the heterogeneous graph, resulting in a surrounding utterance-aware representation for each node.
Finally, we inference the relations between each argument pair based on these coreference-enhanced features.

The coreference-enhanced graph contains four types of nodes and five types of edges where mention nodes, mention-utterance (MU) edges and coreference-chain (CC) edges are proposed based on DialogRE$^{C+}$. 1) Mention nodes are composed of elements in the coreference chain. 2) The MU edge is established to connect a mention node and an utterance node, if the mention is occured in the utterance. 3) In order to establish the interaction among mentions, we fully connect the mention nodes in the coreference chain using CC edges.

\vspace{-2mm}
\subsubsection{REDialog$^{C+}$}
~\cite{zhou2021relation} receives an dialogue with coreference chains as input. First, we use speaker embeddings to represent the speakers of sentences and concatenate them with word embeddings obtained from BERT \cite{devlin2019bert}. Then, we construct a fully-connected coreference-enhanced graph using the feature representation. Finally, the argument nodes after GCN are used for classification.

As shown in Figure~\ref{fig:DRE_graph}.(b), the graph contains three types of nodes. 1) Each mention node corresponds to one mention in the coreference chain of the arguments. The coreference chains of REDialog only contain rule-based pronouns "I" and "you", while our coreference chains also contain rich coreference information such as third person pronouns and possessor pronouns. 2) Argument nodes are the average representations of the corresponding mention nodes. 3) MDP indicates a set of shortest dependency paths for the target entity mentions, and tokens in the MDP are extracted as MDP nodes.

\vspace{-2mm}
\subsubsection{GAIN$^{C+}$}
constructs two graphs based on \cite{zeng2020double}. As shown in Figure~\ref{fig:DRE_graph}.(c), the coreference-enhanced graph contains dialogue node which aims to model the overall dialogue information, and mention nodes denoting each mention of arguments. The introduction of coreference chains enriches the mention nodes in the graph, helping to capture the arguments features distributed in dialogue. The entity-level graph aggregates mentions for the same entities in the former graph.

\vspace{-2mm}
\subsubsection{HGAT$^{C+}$}
~\cite{chen2023dialogue} constructs a graph that contains five types of nodes: argument nodes, utterance nodes, speaker nodes, type nodes and word nodes, where speaker nodes represent each unique speaker in the input dialogue, type nodes denote the word types like PERSON and LOCATION and word nodes denote the vocabulary of dialogue, as shown in Figure~\ref{fig:DRE_graph}.(d).
It also contains eight types of edges, utterance-word/argument/speaker edge, type-word/argument edge and coreference-word(CW)/utterance/(CU)/speaker(CS) edge. 
Each word or argument is connected with the utterances that contain the word or argument, and it is also connected with corresponding types. Each speaker is connected with the utterances uttered by the speaker. 

Due to the introduction of coreference chains, three types of edges have been added to the graph. The CW edges connect the argument and the words in its coreference chain. The CS edges will connect the argument and the same speaker if the argument is a speaker. We have also added CU edges to connect the argument and the utterances where the mentions of this argument occur.

\vspace{-2mm}
\subsection{Coreference Resolution Model}


We use E2E-coref \cite{lee2017end}, the first end-to-end coreference resolution model, to explore the impact of automatic extraction of coreference chains on the DER task.
We train the E2E-coref model using the coreference data from the CoNLL-2012 task \cite{pradhan2012conll} and our DialogRE$^{C+}$ dataset, respectively.
It formulate the coreference resolution task as a set of decisions for every possible span in the document.
For the $i$-th span, the task involves predicting its antecedent $y_i$ based on the coreference score.
The set of possible assignments for each $y_i$ is $\{\epsilon,1,\dots,i-1\}$, where $\epsilon$ is a dummy antecedent. 
First, we obtain the span representations $g$ through pretrained word embeddings and BiLSTM. 
Then, the mention score $s_m$ and antecedent score $s_a$ are calculated based on span representations:
\begin{equation}
    s_m(i)=w_m \cdot FFNN_m(g_i)
\end{equation}
\begin{equation}
    s_a(i,j)=w_a \cdot FFNN_a([g_i,g_j,g_i\circ g_j,\phi(i,j)])
\end{equation}
where $w_m$ and $w_a$ are trainable parameters, $FFNN$ denotes a feed-forward neural network, $\cdot$ denotes the dot product, $\circ$ denotes element-wise multiplication, and $\phi(i,j)$ denotes the feature vector from the metadata. 
The coreference score $s(i,j)$ between span $i$ and span $j$ can be obtained from $s_m$ and $s_a$:
\begin{equation}
    s(i,j)=s_m(i)+s_m(j)+s_a(i,j)
\end{equation}
For more technical details, we recommend referring to the original paper~\cite{lee2017end}.

\vspace{-3mm}
\section{Experiments and Analyses}
\vspace{-3mm}
\subsection{Experiment Setup}
\vspace{-3mm}
We conduct experiments on DialogRE and DialogRE$^{C+}$ and calculate F1 scores as evaluation metrics. 
For all the BERT representations, we use BERT-base-uncased model (768d) as the encoder. 
For GAIN-GloVe, we use the GloVe embedding (100d) and BiLSTM (256d) as word embedding and encoder. 
For HGAT, we use the GloVe embedding (300d) and BiLSTM (128d) as word embedding and encoder.
We set the learning rates of TUCORE-GCN, REDialog, GAIN-BERT GAIN-GloVe, and HGAT to 3e-5, 1e-5, 1e-3, 1e-3, and 1e-4, respectively.

\begin{table*}[!t]
 \setlength{\tabcolsep}{1.2mm}
\centering
\scriptsize
\caption{Model performance comparison. We run each experiment five times and report the average F1 along with standard deviation ($\sigma$).}
\scalebox{0.97}{
\begin{tabular}{lcccccccc}
\hline
\multicolumn{1}{c}{\multirow{2}{*}{\textbf{Model}}} & \multicolumn{4}{c}{\textbf{Dev}}        & \multicolumn{4}{c}{\textbf{Test}}       \\
    \cmidrule(r){2-5}\cmidrule(r){6-9}
  & \multicolumn{1}{c}{\textbf{ori($\sigma$)} }    & \textbf{+cof($\sigma$)}   & \textbf{+ret($\sigma$)}  & \textbf{+zs($\sigma$)}  
  & \multicolumn{1}{c}{\textbf{ori($\sigma$)} }    & \textbf{+cof($\sigma$)}   & \textbf{+ret($\sigma$)}  & \textbf{+zs($\sigma$)}\\
  \hline
TUCORE    & 66.8(0.7)  & 68.8(0.3)  & 67.9(0.6)    & 67.5(0.5)   & 65.5(0.4)   & 67.8(0.4)  & 67.0(0.9)    & 65.8(1.0) \\
REDialog   & 63.0(0.4)  & 65.6(1.1)  & 63.8(0.8)    & 61.2(0.3)   & 62.8(1.5)  & 65.2(1.1)  & 62.5(0.6)    & 59.8(0.3) \\
GAIN$_{BERT}$  & 66.1(0.4)  & 68.4(0.4)  & 65.1(0.3)    & 63.5(0.3)   & 63.6(0.8)  & 67.2(0.6)  & 63.7(0.3)    & 61.6(0.7) \\
GAIN$_{GloVe}$  & 52.3(0.9)  & 57.2(1.0)  & 55.3(0.5)    & 53.3(0.8)   & 55.8(0.3)  & 58.2(0.4)  & 56.3(0.5)    & 55.8(0.4) \\
HGAT   & 56.7(0.3)  & 59.0(0.3)  & 57.9(0.6)    & 57.2(0.3)   & 59.4(0.7)  & 54.7(0.8)  & 57.9(1.0)    & 53.1(0.7) \\
Ave.    & -  & +2.8  & +1.0    & -0.4   & -  & +3.2  & +0.6    & -1.0 \\
\hline
\end{tabular}}
\label{main_results}
\vspace{-1em}
\end{table*}

\vspace{-2mm}
\subsection{Main Results}
Table~\ref{main_results} reports the evaluation results\footnote{We remove the rule-based person references from original model of REDialog.}.
For each of the base models, we compare the performances of the original model (denoted as \textbf{ori}), the model with manually annotated coreference information (\textbf{+cof}), the model with the coreference information generated from a pre-trained coreference resolution model (\textbf{+ret}), the model with the coreference information parsed from an off-the-shelf coreference resolution tool \cite{lee2017end} (namely zero-shot, \textbf{+zs}).
From Table~\ref{main_results}, we have the following observations:
1) The annotated coreference chains introduce stable F1 improvements for all base models, with an average improvement of 2.8 and 3.2 on dev set and test set respectively. This validates the effectiveness of coreference chains on graph-based models in relational reasoning.
2) The retrained coreference chains achieve a lower improvement effect than the the annotated ones. This demonstrates the practicality of coreference chains in relation extraction.
3) The zero-shot coreference chains do not show any improvement effect on DRE, indicating the necessity of annotating DialogRE$^{C+}$ dataset.

\begin{table}[!th]
\centering%
    \begin{minipage}[t]{0.4\linewidth}
        \centering
        \caption{Ablation studies of different edge types.}
        \begin{tabular}{lcc}
        \hline
        \textbf{Model}  & \textbf{dev}   & \textbf{test}  \\ 
        \hline
        TUCORE-GCN$^{c+}$  & \bf 68.8 & \bf 67.8 \\
        \hline
        \quad w/o CC edge        & 67.4 & 66.8 \\
        \quad w/o MU edge        & 67.9 & 67.5 \\
        \quad w/o CC and MU edge & 66.8 & 65.5 \\ \hline
        \end{tabular}
        \label{table:ablation}
        \vspace{-1em}
    \end{minipage}%
    \hspace{.1in}
    \begin{minipage}[t]{0.55\linewidth}
        \renewcommand\arraystretch{1.75}
        \scriptsize
        \caption{Performance comparison of inter- and intra-utterance relation extraction.}
        \centering
        \begin{tabular}{lcccc} 
        \hline
        \multicolumn{1}{c}{\multirow{2}{*}{\small\textbf{Method}}}& 
        \multicolumn{2}{c}{\small\textbf{dev}} & 
        \multicolumn{2}{c}{\small\textbf{test}} \\ \cline{2-5} 
        \multicolumn{1}{c}{} & \small{Inter} & \small{Intra} & \small{Inter} & \small{Intra} \\ \hline
        \small{TUCORE-GCN$^{c+}$} & \small{65.9} & \small{65.4} & \small{66.3} & \small{63.8}  \\
        \small{TUCORE-GCN}     & \small{65.3} & \small{63.7} & \small{65.2} & \small{63.5}       \\ \hline
        \end{tabular}
        \label{table:cross}
        \vspace{-1em}
    \end{minipage}
\vspace{-2mm}
\end{table}

\vspace{-3mm}
\subsection{Effect of Coreference-Enhanced Graphs}
We conducted ablation experiments based on TUCORE-GCN$^{c+}$. 
We remove the coreference-chain and mention-utterance edges respectively and simultaneously.
The coreference-chain edges are removed to judge whether the coreference information has the ability to learn from each other, the mention-utterance edges are removed to prove the effect of location information of coference chain. The graph structure of TUCORE-GCN is formed by removing  both the coreference-chain edges and mention-utterance edges.
Table~\ref{table:ablation} shows the result of the above three experiments, the names of edges are capitalized, for example, coreference-chain edge is represented as CC edge. The result shows that the use of coreference chains to enrich the dialogue graph is beneficial for relationship extraction.

\begin{figure}[!t]
    \begin{minipage}[t]{0.55\linewidth}
        \centering
        \includegraphics[width=\textwidth]{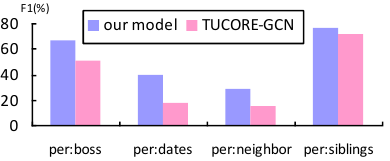}
        \caption{Effect analysis for partial relations.}
        \label{fig:partial}
    \end{minipage}%
    \hspace{.1in}
    \begin{minipage}[t]{0.4\linewidth}
        \centering
        \includegraphics[width=\textwidth]{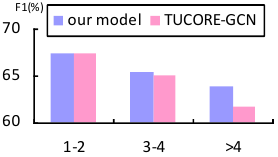}
        \caption{Effect analysis for different speaker numbers.
        }
        \label{fig:sk-nums}
    \end{minipage}
\vspace{-2mm}
\end{figure}

\vspace{-3mm}
\subsection{Effect Analysis for Partial Relationships}
In Figure~\ref{fig:partial}, we compare the performances of TUCORE-GCN$^{c+}$ and TUCORE-GCN in extracting partial relationships.
Obviously, in those relationships, such as per:boss and per:neighbor, the prediction effect of the model is significantly improved. 
The reason may be that in the communication, the speaker will add the possessive pronouns of sequential adjectives when referring to this kind of relationship, such as `your, his, her', when joining the labeled pronouns, the model performs well in the prediction process.

\vspace{-3mm}
\subsection{Impact of Coreference on Speaker Numbers}
In the Figure~\ref{fig:sk-nums}, we compare the performance of TUCORE-GCN$^{c+}$ and TUCORE-GCN in the dialogues with different speaker numbers. When the number of speakers is greater than 4, the effect of the model is improved after adding coreferential information. This is because when there are a large number of speaker, the dialogue is generally longer, and the prediction effect in the longer text can be improved by adding coreference information.

\vspace{-3mm}
\subsection{Impact of Coreference on Inter- and Intra-Utterance Relations}
In this section, we make a comparative analysis for inter- and intra-utterance RE using TUCORE-GCN$^{c+}$ and TUCORE-GCN.
In Table~\ref{table:cross}, we can find that in the dev and test datasets, the F1 of inter-utterance relation extraction increased by 0.6 and 1.1, which shows that joining coreference information can promote cross-utterance relationship extraction effect, which is due to join mention node in different utterances and join different mention in the same utterance, thus the model can learn more information in different utterances more accurately, so as to improve the effect of inter-utterance relation extraction.

\vspace{-3mm}
\section{Conclusion and Future Work}
\vspace{-3mm}
In this paper, we annotate coreference information based on the DialogRE dataset, proposing the first dialogue relation extraction dataset with coreference chains. Based on this dataset, we build 4 coreference-enhanced graph-based models for DRE. By adding annotated coreference information into DRE models, the effects of the models exceed the baselines significantly. 
We also add automatically extracted coreferentce chains into DRE models. The coreferentce chains given by a retrained coreference resolution model help DRE models to some extent, while the ones given by an off-the-shelf coreference resolution model worsen the performances of DRE models, demonstrating the necessity of annotating coreferentce chains on the DRE task.

In the following research work, we hope to make contributions to the identification of dialogue coreference information based on the proposed DialogRE$^{C+}$ dataset, use coreference chains to improve DRE and apply the experimental results to other dialogue-level tasks. 
We are also interested in joint research on the DRE task and coreference resolution task.

\vspace{-3mm}
\subsubsection*{Acknowledgement.}
This work is supported by the National Key Research and Development Program of China (No. 2022YFB3103602) and the National Natural Science Foundation of China (No. 62176187).

%
%
%
\bibliographystyle{splncs04}
\bibliography{mybibliography}

\end{CJK}
\end{document}